\pdfoutput=1

\documentclass[11pt]{article}

\usepackage[]{EMNLP2023}

\usepackage{times}
\usepackage{latexsym}

\usepackage[T1]{fontenc}

\usepackage[utf8]{inputenc}

\usepackage{microtype}
\usepackage{graphicx}
\usepackage{amsmath}
\usepackage{amsthm}
\usepackage{booktabs}
\usepackage{algorithm}
\usepackage{algorithmic}
\usepackage{mathtools}
\usepackage{booktabs}
\usepackage{csquotes}
\usepackage{multirow}
\usepackage{amssymb}
\usepackage{wasysym}

\usepackage{inconsolata}

\title{Weakly-supervised Deep Cognate Detection Framework for Low-Resourced Languages Using Morphological Knowledge of Closely-Related Languages}

\author{
Koustava Goswami$^{1,2}$\thanks{$^*$Research work conducted during his time at the Data Science Institute, University of Galway, Ireland. Contact: koustavag@adobe.com}~~, Priya Rani$^2$, Theodorus Fransen$^{2,3}$, \\ {\bf John P. McCrae}$^2$ \\
$^1$ Adobe Research Bangalore, India \\
$^2$  Data Science Institute, University of Galway, Ireland\\
$^3$  Università Cattolica del Sacro Cuore, Milan, Italy\\
{\tt \small koustavag@adobe.com},
{\tt \small \{p.rani1,john.mccrae\}@universityofgalway.ie}, {\tt \small theodorus.fransen@unicatt.it}
}

\begin{document}
\maketitle

\begin{abstract}
Exploiting cognates for transfer learning in under-resourced languages is an exciting opportunity for language understanding tasks,
including unsupervised machine translation, named entity recognition and information retrieval. Previous approaches mainly focused on supervised cognate detection tasks based on orthographic, phonetic or state-of-the-art contextual language models, which under-perform for most under-resourced languages. This paper proposes a novel language-agnostic weakly-supervised deep cognate detection framework for under-resourced languages using morphological knowledge from closely related languages. We train an encoder to gain morphological knowledge of a language and transfer the knowledge to perform unsupervised and weakly-supervised cognate detection tasks with and without the pivot language for the closely-related languages. While unsupervised, it overcomes the need for hand-crafted annotation of cognates. We performed experiments on different published cognate detection datasets across language families and observed not only significant improvement over the state-of-the-art but also our method outperformed the state-of-the-art supervised and unsupervised methods.  Our model can be extended to a wide range of languages from any language family as it overcomes the requirement of the annotation of the cognate pairs for training. The code and dataset building scripts can be found at \url{https://github.com/koustavagoswami/Weakly_supervised-Cognate_Detection}
\end{abstract}

\section{Introduction}


Cognates are etymologically related word pairs across languages \cite{crystal2011dictionary}. 
However, cognates are defined in much broader terms in many different fields, including natural language processing (NLP) or psycholinguistics \cite{labat-lefever-2019-classification}. In these areas, word pairs with similar meanings and spelling are also considered as cognates. In the recent development of automatic machine translation (AMT), automatic cognate detection is found to be very effective for similar language translation \cite{kondrak2005cognates,DBLP:conf/naacl/KondrakMK03}. Moreover, it helps to efficiently perform cross-lingual information retrieval \cite{makin2007experiments,meng2001generating} from different sources. Very often, words that have similar spelling are recognised as cognates (Example: the Latin and the English word pair \enquote{cultūra} and \enquote{culture}). Nevertheless, there are word pairs which are false friends or partial cognates \cite{DBLP:conf/lrec/KanojiaKBH20}. Partial cognates are similar words across languages but carry different meanings in different contexts \cite{kanojia2019utilizing}, thus making automatic cognate detection hard and challenging. Identifying these cognates requires extensive linguistic knowledge across languages, which is quite hard and expensive to annotate. While cross-lingual automatic cognate detection systems exist, they have primarily been supervised methods requiring labelled data or language-specific linguistic rules. For under-resourced languages, finding annotators or linguists is a challenging task. This highlights the need for an efficient unsupervised language-agnostic cognate detection framework. We show that our weakly-supervised and unsupervised approaches can better exploit the available data than existing supervised methods and thus produce better results for under-resourced languages. The method transfers the morphological knowledge of a shared encoder in the unsupervised cognate detection framework into a Siamese network setting, where the framework simultaneously learns word representation and cluster assignments in a self-learning setup.  

Supervised cognate detection frameworks understand the relationship between word pairs by concentrating on their phonetic or lexical similarities based on their annotated positive or negative labels \cite{jager-etal-2017-using,DBLP:conf/coling/Rama16}. Recently, researchers tried to exploit contextual multilingual word embedding techniques to identify cognates which produced better results than only concentrating on phonetic transcriptions \cite{DBLP:conf/coling/KanojiaDDBHK20}. Although such methods had good results, annotating labels is quite expensive and tedious for many under-resourced languages. Moreover, producing multilingual contextual word embeddings is challenging for these languages due to the need for more data sources on the web. \citet{DBLP:conf/conll/MerloR19} highlighted that based on bilingual lexicon matching between two known languages, the similarity score produced by contextual word embeddings could differentiate between true or false cognate pairs. Though this technique is label independent, the framework depends on the bilingual lexicon availability of the known language pairs.

To alleviate the above challenges, in this paper, we propose a {\em language-agnostic weakly-supervised cognate detection framework based on Siamese architecture} with an iterative clustering approach \cite{xie2016unsupervised} during back-propagation. Our encoder design is inspired by \citet{DBLP:conf/coling/GoswamiSCFM20}, where they learn the n-gram character features of a sentence with attention. We introduce a {\em positional encoder} on n-gram features, which, in combination with the attention mechanism, learns sub-word representations of a word. We also depict the performance of our {\em morphological knowledge-based weakly-supervised framework}. This variant gives a better understanding of the grammar and structural analysis of the words of a language. Thus, transferring this knowledge with the help of a shared encoder to closely-related languages enhances the understanding of structural and grammatical relatedness between cross-lingual word pairs. Moreover, our word encoding method helps to produce better supervised cognate detection results, which outperform the state-of-the-art supervised results on various language pairs. In this paper, we have presented a complete set up of results for supervised, weakly-supervised and unsupervised setups.

The extensive experiments (in Section \ref{sec:exper}) on three different cognate detection datasets across language families have showcased the efficacy of our weakly-supervised and supervised cognate detection framework. For example, on six different Indian language pairs, our weakly-supervised model (with morphological knowledge) has outperformed the state-of-the-art supervised model proposed by \citet{DBLP:conf/coling/KanojiaDDBHK20} by an average of ${\bf 9}$ points of $F$-score whereas, for Celtic language pairs, it outperformed by ${\bf 8.6}$ points of $F$-score. At the same time, our supervised framework has produced a state-of-the-art performance by outperforming the existing supervised model by an average of ${\bf 16}$ points of $F$-score. Thus, our model is robust across diverse language families for the supervised and weakly-supervised cognate detection task. Interestingly, our experiments show that on Indian language pairs like Hindi-Punjabi and Hindi-Marathi, an encoder with morphological knowledge of the Hindi language performed better than an encoder with morphological knowledge of their ancestral language, Sanskrit by an average of ${\bf 1.5}$ points of $F$-score. However, the performance of the weakly-supervised framework for the language pair Marathi-Bengali has improved by ${\bf 2}$ points of $F$-score. 

In a nutshell, our contributions are:
\begin{itemize}
    \item {\bf (i)} a language-agnostic weakly-supervised cognate detection framework without the need for labels,
    \item {\bf (ii)} efficiently transferring morphological knowledge of a low-resourced language to closely-related under-resourced languages with or without the need for the pivot language for better cognate detection,
    \item {\bf (iii)} introduction of positional embedding along with attention to different n-grams of a word for better understanding of word structures,
    \item {\bf (iv)} robustness in weakly-supervised and supervised cognate detection for low-resource languages across three different datasets of different language families, outperforming state-of-the-art supervised approaches.
\end{itemize}

\section{Related Work}
\label{sec:relat}

Algorithms for automatic cognate detection (ACD) are mostly based on phonetic or orthographic similarity measures and often language-dependent or supervised approaches. \citet{covington1996algorithm} developed an algorithm to align historical-comparative languages by their phonetic similarity. \citet{kondrak2000new} released a novel algorithm for aligning phonetically similar sequences. Similarity-based algorithms follow these works to identify cognates between a language pair of different families. The orthographic similarity-based works calculate distances or string similarities between word pairs and define similarity scores to identify cognates   \cite{mulloni-pekar-2006-automatic,melamed-1999-bitext,jager-etal-2017-using}. \citet{DBLP:conf/coling/Rama16} has released a convolution-based model that also considers phonetic similarity-based scores into account to detect cognates for word pairs. Some researchers have also taken parallel datasets into account to identify cognates. Distance measurement based scores have become the feature set to identify cognates in these cases \cite{mann-yarowsky-2001-multipath,tiedemann-1999-automatic}. \citet{10.1145/3297001.3297045} performed a cognate detection task on Indian languages, which includes a large amount of manual intervention during identification. \citet{kanojia2019utilizing} introduced a character sequence-based recurrent neural network for identifying cognates between Indian language pairs.

The influence of classical machine learning and dynamic programming-based approaches defines automatic cognate detection tasks as semi-supervised approaches. \citet{hauer2011clustering} trained a linear SVM based on word similarity and language-pair features to detect cognates. Phonetic alignment based SVM models performed quite efficiently on different language families while detecting cognates \cite{jager-etal-2017-using}. Some researchers designed orthographic substring similarity measures based SVM models \cite{ciobanu2014automatic,ciobanu2015automatic}. 

Another thread of research for cognate detection recently involved multilingual contextual knowledge injection methods. \citet{merlo2019cross} explored the effect of cross-lingual features on bilingual lexicon building, which was later implemented in Indian language cognate detection methods. \citet{DBLP:conf/coling/KanojiaDDBHK20} injected cross-lingual semantic features for cognate detection tasks using newly trained language models, which showed better results than state-of-the-art methods. Recently, \citet{kanojia-etal-2021-cognition} proposed incorporating gaze features in context aware cognate detection tasks, which improved results for Hindi-Marathi language pairs. These methods may produce better results; however, the main disadvantages lie in training these models as they are data-hungry models. Thus, incorporating manually annotated gaze features and newly generated cross-lingual contextual embeddings becomes an impossible solution for many under-resourced languages.

\section{Cognate Detection Framework}
\label{sec:model}

In this section, we describe the components and working of the proposed {\em Language Agnostic Weakly-supervised Cognate Detection Architecture Using Morphological Knowledge}, trained using an unsupervised loss function and iterative clustering method. The iterative clustering process enhances the understanding of word representation and cluster assignment. 

\begin{figure}[t]
    \centering
    \includegraphics[width=\columnwidth]{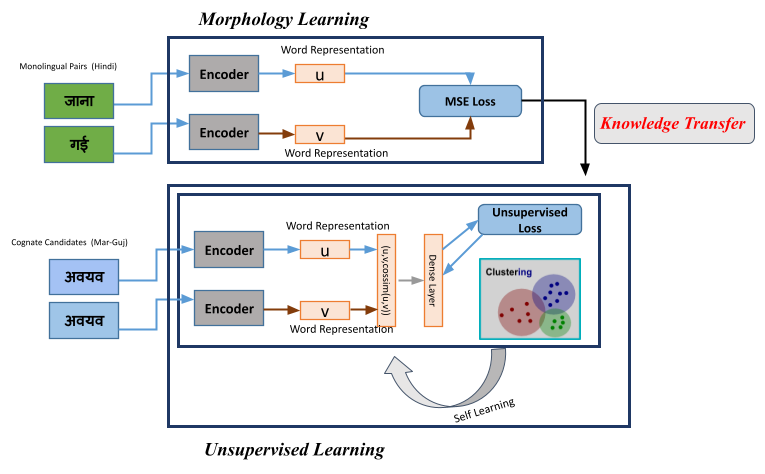}
    \caption{Weakly-supervised Cognate Detection Framework with Morphology Learner and Unsupervised Cognate Detector. For training, we pass monolingual word-pairs into morphology learners (coloured in green) and bilingual cognate candidates (coloured in blue) into unsupervised cognate detector.}
    \label{fig:unsup_framework}
\end{figure}

Figure \ref{fig:unsup_framework} depicts the shared word encoder based framework. It consists of two parts - {\bf (i)} a {\em Morphology Learner}, which gathers the morphological knowledge of a language and {\bf (ii)} an {\em Unsupervised Cognate Detector}, which uses the morphological knowledge learnt using a shared word encoder to cluster the cognates between word pairs.

\subsection{Word Encoder}

 The word encoder consists of an n-gram character-level CNN and a positional embedding layer, followed by a self-attention layer.
 
 {\bf Character Encoding.}  The character level CNN layer generates word representation on n-gram ($n\in\left\{\mbox{2,3,4,5,6}\right\}$) characters, which helps to understand the representation of words on different sub-word levels \cite{DBLP:conf/coling/GoswamiSCFM20}. A word's different sub-word level representations are achieved using a 1-dimensional CNN \cite{zhang2015character}. The characters of a word are fed as input sequence $S = [w_1,w_2,...,w_m]$ to the 1-dimensional CNN, where $m$ is the number of characters present in the word and $w_i$ is a character in the word. Considering the 1-dimensional CNN as a feature extractor, it slides over characters to create a window vector $w_j$ with consecutive character vectors, as denoted in Equation \ref{equation-1}.
 
 \begin{equation}\label{equation-1}
    w_j=[x_j,x_{j+1},.....,x_{j+k-1}]
\end{equation}
where \textit{k} is the size of the feature extractor filter and $x_i\in\mathbb{R}^d$ is the d-dimensional character embedding of the \textit{i}-th character, where character vocabulary size is $n$. Thus this $k$-sized filters create the feature map $s$ ($s\in\mathbb{R}^{m-k+1} $) from the window vector $w_j$ according to Equation \ref{equation-2},

\begin{equation}\label{equation-2}
    s_j = a(w_j \cdot m + b_j)
\end{equation}
where the vector \textit{m} is a filter for convolution operation, $b_j$ is the bias for the \textit{j}-th position and \textit{a} is the non-linear function. Thus the new feature representation of the word $F$ ($F\in\mathbb{R}^{(m-k+1) \times n}$) will be expressed as  $F = [s_1,s_2,...,s_n]$, where n is the number of filters and \textit{m} is the total input size. Observe the different filter size $k\in\left\{\mbox{2,3,4,5,6}\right\}$, representing the word's n-gram features, further represented as $F_k$.

{\bf Positional Encoding of Features.} While getting n-gram features of the word representation, the character sequence order carries a significant role in word construction. We try to learn the different n-gram positions in a word with the help of our new introduction of positional encoding. The Transformer architecture \cite{DBLP:conf/nips/VaswaniSPUJGKP17} enforces the trainable positional embeddings on the input word pieces to understand the position of the words in a sentence. In our input words, the same character can appear in multiple positions, which makes positional embedding on each character irrelevant.  Rather, our approach to learning the n-gram sequence in a word helps to understand the grammatical and morphological differences from the structural perspective of a word. Following the learning process of the positional embedding of transformer architecture, we encode the trainable positional encoding of different n-gram features.

{\bf Attention of Features.} The new encoded feature representation $F_k$ produces the ultimate word embedding, which is achieved by giving weight to n-grams according to their importance in word construction. A self-attention mechanism takes the feature representation as input and produces an output weight vector $a$ for every feature representation $F$ using the following Equation \ref{equation-softmax}.

\begin{equation}\label{equation-softmax}
    a = \text{softmax}(\tanh(W_h\cdot F^T + b_h))
\end{equation}
The summation of feature representation $F$ according to the weight vectors provided by $a$ generates a vector representation $r$ of a word by Equation \ref{attention}

\begin{equation}\label{attention}
    r = \sum_{i=1}^{T}a_i \cdot F_i 
\end{equation}
where $a_i$ represents the attention weights, and $\cdot$ represents the element-wise product between elements. The final vector representation $r$ is the concatenation of different n-grams $\in\left\{\mbox{2,3,4,5,6}\right\}$, which is represented as $r=[r_2,r_3,r_4,r_5,r_6]$.

\subsection{Morphology learner}

We learn the morphological relationship between two words \textit{$r_i$} and \textit{$r_j$} of the same language in a Siamese setting (details of the morphological training dataset building with an example are given in Section \ref{sec:train}). The encoded vector representations are then passed through a  fully connected (FC) layer which gives two vector representations $z_l\in\mathbb{R}^{N\times K}$ and $z_r\in\mathbb{R}^{N\times K}$. The morphological learner model is then trained to minimize the mean-squared loss between two word representations such that their vector space distances reflect their degree of morphological relatedness in Equation \ref{mse}

\begin{equation}\label{mse}
    1/N\sum_{i=1}^{N}(z_{l_i} - z_{r_i})^2
\end{equation}
where \textit{$N$} is the mini-batch size.

\subsection{Weakly-supervised/Unsupervised Cognate Detector}

The encoder with and without morphological knowledge for weakly-supervised and unsupervised methodology respectively accepts two word representations \textit{$r_i$} and \textit{$r_j$} from two different languages as input in a Siamese setting. The encoded vector representations are then passed through a  fully connected (FC) layer which gives two vector representations $u\in\mathbb{R}^{N\times K}$ and $v\in\mathbb{R}^{N\times K}$. We concatenate the word representations $u$ and $v$ with their cosine similarity score and pass it through a sense layer to achieve the combined representation $z\in\mathbb{R}^{N\times K}$. It is then passed through a softmax layer to get the probability distribution of all classes $p\in\mathbb{R}^{N\times K}$,  as per Equation \ref{soft}    

\begin{equation}\label{soft}
    p_{ij} = \frac{exp(z_{ij})}{\sum_{t=1}^{K}exp(z_{ik})}
\end{equation}
where \textit{$k$} is the number of classes. We train the unsupervised model based on the maximum likelihood clustering loss proposed by \citet{DBLP:conf/coling/GoswamiSCFM20}, where they try to maximize the probability distribution function for each class and at the same time try to minimize the probability of all the datasets to be assigned to one class using Equation \ref{lossfunction}. 

\begin{equation} \label{lossfunction}
    L_u = \sum_{i=1}^{N}\max_{j=1}^{i}p_{ij} - \max_{i=1}^{N}\sum_{j=1}^{i}p_{ij}^2
\end{equation}

While the unsupervised loss function helps us to get word embeddings and an initial cluster assignment, it is important to improve cluster purity according to datasets. \citet{xie2016unsupervised} proposed a self learning based deep clustering technique. The framework learns the clustering based on stochastic gradient descent (SGD) during backpropagation. We fine-tune our word embedding to learn better clustering using this iterative clustering technique. The initial sets of cluster centroids ${{u_j}}_{j=1}^k$ are obtained from the pre-training phase using Equation \ref{lossfunction}. In this self training phase, we assign word embeddings to initial cluster centroid and then fine-tune the word embeddings and cluster centroids using auxiliary target distribution.  

The assignment of cluster centroids (\textit{$u_j$}) and word embeddings (\textit{$z_i$}) are calculated based on  Student’s t-distribution \cite{maaten2008visualizing} as per Equation \ref{t-dis}

\begin{equation}\label{t-dis}
    q_{ij} = \frac{\left ( 1 + \left \| z_i - u_j \right \|^2 \right )^{-1}}{\sum_{j'}\left ( 1 + \left \| z_i - u_{j'} \right \|^2 \right )^{-1}}
\end{equation}
where $q_{ij}$ is the probability of sample \textit{i} to cluster \textit{j} assignment.

We now  refine the cluster learning from their high confidence assignments using auxiliary target distribution $p_{ij}$ \cite{xie2016unsupervised}. This helps to improve cluster purity by putting more emphasis on data point assignment, as per Equation \ref{prob}

\begin{equation} \label{prob}
    p_{ij} = \frac{q_{ij}^2 / \sum_i q_{ij}}{\sum_{j'}\left (q_{ij}^2 / \sum_i q_{i{j'}}  \right )}
\end{equation}
where $\sum_i q_{ij}$ is the frequency of clusters.

The cluster assignment self-learning process is trained based on KL divergence loss between assignments $q_i$ and $p_i$, as shown in Equation \ref{kl}.
 
\begin{equation} \label{kl}
    L = KL\left ( P || Q \right ) = \sum_i \sum_j p_{ij} \log \frac{p_{ij}}{q_{ij}}  
\end{equation}

\section{Training and Evaluation Dataset}
\label{sec:train}

\begin{table}[t]
\centering
\small
	\begin{tabular}{|l|l|}
	\hline
	{\bf Word1} & {\bf Word2}  \\
	\hline
	nuachtán. & nuachtáin  \\
	\hline
	eolas. & a eolais. \\
	\hline
	síceolaí. & leis an síceolaí.  \\
	\hline
	Críostaí. &  Críostaithe.  \\
	\hline
	\end{tabular}
	\caption{Morphology Learning Dataset for the Irish.}
	\label{tab:data_set}
\end{table}

The framework has two parts: {\bf(i)} Morphology Learner and {\bf(ii)} Weakly-supervised/Unsupervised Cognate Detector. 

{\bf Morphological training} is based on UniMorph \cite{mccarthy-etal-2020-unimorph} datasets. As shown in Table \ref{tab:data_set}, our Siamese network accepts two words as input. The inputs are the monolingual word pairs of the pivot language in UniMorph data (as shown in Figure \ref{fig:unsup_framework}, we train the encoder of the morphological learner on the Hindi dataset in a supervised manner and transfer the knowledge to the unsupervised cognate detector for Marathi-Gujrati word pairs). Though our model is trained with the supervised dataset from Unimorph, we do not consider the annotated morphological class while training the morphological learner. The statistics of datasets for three pivot languages and Sanskrit are given in Table \ref{tab:morph_data}.

\begin{table}[t]
\centering
\small
\scriptsize

\resizebox{\columnwidth}{!}{%
	\begin{tabular}{|l|l|l|l|l|}
	\hline
	{\bf Dataset} & {\bf Hindi} & {\bf Irish} & {\bf Zulu} & {\bf Sanskrit}\\
	\hline
	Training Dataset. & 42200 & 2579 & 49696 & 437675 \\
	\hline
	\end{tabular}
}
	\caption{Morphological Training Dataset of Three Pivot Languages and Sanskrit}
	\label{tab:morph_data}
\end{table}

{\bf Cognate Detection Task}  We have evaluated our models on three different datasets for the cognate detection task: Indian, Celtic, and South African languages. For under-resourced Indian language pairs, we have followed the work of \citet{DBLP:conf/lrec/KanojiaKBH20}.  As datasets for South African and Celtic languages are not easily available, we built the dataset from an open-source cognate database \cite{batsuren-etal-2019-cognet} and also have used the SigTyP 2023 shared task on cognate detection dataset \cite{rani-etal-2023-findings}. The true cognates are directly taken from the dataset and false cognate pairs are randomly shuffled word-pairs with a $60$-$40$ split of the total dataset available in the database for each language pair. We experiment with both supervised and unsupervised learning of these cognate classifiers based on the encoder that was learned in the previous step. During the training procedure of the unsupervised cognate detector, no word pair labels of cognate datasets are considered. The detailed statistics of the cognate datasets for the three language families can be found in Table \ref{tab:test_data_set}.

\begin{table}[t]
\centering
\scriptsize
	\begin{tabular}{|l|l|l|}
	\hline
	{\bf Language-pairs} & {\bf Cognates} & {\bf Non-cognates} \\
	\hline
	Hindi (Hi) - Marathi (Mr) & 15726  & 15983  \\
	\hline
	Hindi (Hi) - Gujrati (Gu) & 17021 & 15057  \\
	\hline
	Hindi (Hi) - Punjabi (Pa) & 14097 & 15166  \\
	\hline
	Hindi (Hi) - Bengali (Ba) &  15312 & 16119  \\
	\hline
	Hindi (Hi) - Tamil (Ta) &  3363 & 4005  \\
	\hline
	Hindi (Hi) - Assamese (As) &  3478 & 4101  \\
	\hline
	Irish (Ga) - Manx (Gv) &  335 & 223  \\
	\hline
	Irish (Ga) - Scottish Gaelic (Gd) &  676 & 450  \\
	\hline
	Zulu (Zu) - Xhosa (Xh) &  2236 & 1490  \\
	\hline
	Zulu (Zu) - Swati (Ss) &  14 & 9  \\
	\hline
	\end{tabular}
	\caption{Cognate dataset statistics across language-pairs.}
	\label{tab:test_data_set}
\end{table}

For all of our experiments we have carried out \textit{$5$-fold stratified cross-validation} which has helped us to get the train and test data randomly. 

\section{Training Details}

We implemented our model using pytorch\footnote{\url{https://pytorch.org}}. The learning rate for the Indian, Celtic and South African datasets was hand-tuned to 1e-4, 2e-3 and 4e-3, respectively, for the morphological training. At the same time, for the unsupervised cognate detection tasks the learning rates were 1e-2,1e-1,1e-2, respectively. To stabilize the learning of the model, we have implemented LambdaLR\footnote{\url{https://pytorch.org/docs/stable/optim.html}} as the learning rate scheduler. 
For clustering, we have used \textit{k}-means clustering \cite{macqueen1967} with minibatch \footnote{\url{https://scikit-learn.org/stable/modules/generated/sklearn.cluster.MiniBatchKMeans.html}}.

\section{Experimental Evaluation}
\label{sec:exper}

We evaluated our framework on three different datasets in three different scenarios: (a) language pairs with pivot language and its morphological knowledge, (b) language pairs without the pivot language but with the shared encoder having morphological knowledge of the pivot language and (c) the effect of the historical language morphological knowledge transfer on the language pairs. 

We compare our models with the following supervised state-of-the-art cognate detection frameworks: (i) {\bf CNN based model} Siamese CNN based approach by \citet{DBLP:conf/coling/Rama16}, (ii) {\bf Orthographic similarity} based approach from \citet{labat-lefever-2019-classification}, (iii) {\bf Recurrent Neural Network} based approach proposed by \cite{kanojia2019utilizing}, (iv) {\bf Contextual Word Embedding} based approach with XLM-R \cite{DBLP:conf/acl/ConneauKGCWGGOZ20} proposed by \citet{DBLP:conf/coling/KanojiaDDBHK20}.

\begin{table}[t]
\centering
\scriptsize
\begin{tabular}{lp{0.25cm}p{0.3cm}p{0.3cm}p{0.3cm}p{0.3cm}p{0.3cm}}
\toprule
{\textbf{Approaches / Languages}} & {\textbf{Hi-Mr}} & {\textbf{Hi-Gu}} & {\textbf{Hi-Pa}} & {\textbf{Hi-Bn}} & {\textbf{Hi-Ta}} & {\textbf{Hi-As}} \\ \midrule

\midrule
{Orthographic Similarity}    & {0.21}          & {0.23}          & {0.21}          & {0.36}          & {0.20}          & {0.34}          \\
\cite{DBLP:conf/coling/Rama16} & {0.69} & {0.67}          & {0.47} & {0.65} & {0.53} & {0.71} \\ 
\cite{kanojia2019utilizing}                     & {0.72}          & {0.76}          & {0.74}          & {0.68}          & {0.53}          & {0.71}          \\ 
{XLM-R + FFNN}                     & {0.73}          & {0.76}          & {0.73}          & {0.78}          & {0.56}          & {0.71}          \\  \midrule
\multicolumn{7}{c}{\textbf{Proposed Supervised Methods}} \\
\hline
\midrule
{\em Proposed-method}                & {0.81}          & {0.79}          & {0.80}          & {0.79}          & {0.69}          & {0.78}          \\ 
{\em $\text{Proposed-method}_\text{withknowledge}$}                & \textbf{{0.91}}          & \textbf{{0.87}}          & \textbf{{0.88}}          & \textbf{{0.86}}           & \textbf{{0.77}}          & \textbf{{0.82}}          \\ \midrule
\multicolumn{7}{c}{\textbf{Proposed Weakly-supervised/Unsupervised methods}} \\
\hline
\midrule
{\em $\text{Proposed-method}_\text{unspv}$}                 & {0.72} & {0.73} & {0.74} & {0.75} & {0.67} & {0.69} \\ 
{\em $\text{Proposed-method}_\text{wklysupv}$}                & {\textbf{0.85}} & {\textbf{0.84}} & {\textbf{0.81}} & {\textbf{0.82}} & {\textbf{0.74}} & {\textbf{0.79}} \\ \bottomrule
\end{tabular}
\caption{Results of supervised and weakly-supervised cognate detection task based on $\text{F}$-Score for Indian languages. The baseline performances are as reported in~\protect\cite{DBLP:conf/coling/KanojiaDDBHK20}.}
\label{table:1}
\end{table}
\begin{table}[t]
\centering
\scriptsize
\begin{tabular}{lp{0.25cm}p{0.3cm}p{0.3cm}p{0.3cm}}
\toprule
{\textbf{Approaches / Languages}} & {\textbf{Zu-Ss}} & {\textbf{Zu-Xh}} & {\textbf{Ga-Gd}} & {\textbf{Ga-Gv}} \\ \midrule
\midrule
{Orthographic Similarity}    & {0.21}          & {0.31}          & {0.29}          & {0.22}          \\
\cite{DBLP:conf/coling/Rama16} & {0.24} & {0.61}          & {0.64} & {0.59} \\ 
\cite{kanojia2019utilizing}                     & {0.23}          & {0.74}          & {0.72}          & {0.61} \\ \midrule
\multicolumn{5}{c}{\textbf{Proposed Supervised Methods}} \\
\hline
\midrule
{\em Proposed-method}                & {0.65}          & {0.76}          & {0.77}          & {0.69}  \\ 
{\em $\text{Proposed-method}_\text{withknowledge}$}                & \textbf{{0.72}}          & \textbf{{0.87}}          & \textbf{{0.88}}          & \textbf{{0.74}}                     \\ \midrule
\multicolumn{5}{c}{\textbf{Proposed Weakly-supervised/Unsupervised methods}} \\
\hline
\midrule
{\em $\text{Proposed-method}_\text{unspv}$}                & {0.69} & {0.73} & {0.71} & {0.62} \\ 
{\em $\text{Proposed-method}_\text{wklysupv}$}                & {\textbf{0.78}} & {\textbf{0.79}} & {\textbf{0.81}} & {\textbf{0.71}} \\ \bottomrule
\end{tabular}
\caption{Results of supervised and weakly-supervised cognate detection task based on $\text{F}$-Score for South African and Celtic languages.}
\label{table:2}
\end{table}

\subsection{Pivot Language based Cognate Detection}
\label{ssec:sts}

Understanding word representation is the key to state-of-the-art deep learning frameworks for different cross-lingual cognate detection tasks. Supervised models rely on distributional learning based on annotated labels. In contrast, weakly-supervised and unsupervised frameworks should be able to learn the structural and syntactical representations of words to do clustering. We have evaluated our model on different language families, including Indo Aryan, Dravidian \cite{DBLP:conf/lrec/KanojiaKBH20}, Celtic and South-African languages.

As shown in Table \ref{table:1}, we have evaluated our model on six different language pairs. Our baseline model is based on the orthographic similarity approach. As expected, it does not perform well (Example: Word pair \enquote{Alankar (Ornament)} in Hindi, and \enquote{Alankaaram (Ornament)} in Tamil with similar word structures classified wrongly). The character-based CNN method proposed by \citet{DBLP:conf/coling/Rama16} followed by the recurrent network-based solution proposed by \citet{kanojia2019utilizing} increases the model efficiency by quite a margin while detecting cognates. The contextual word embedding XLM-R based baseline model gives the best score compared to the previous models' score. This model is capable of injecting contextual knowledge of words from a sentence. Our proposed approach has outperformed all of these baseline frameworks and achieved state-of-the-art results for supervised and weakly-supervised frameworks. As the language pairs Hindi-Marathi are very closely related, transferring the learnt Hindi morphological knowledge has increased the model's efficiency by ${\bf 18}$ points of $F$-score. We observed that our weakly-supervised framework outperformed the state-of-the-art supervised baseline system by ${\bf13}$ points of $F$-score. It is interesting to note that our supervised and unsupervised frameworks achieved state-of-the-art results by outperforming the baseline systems by ${\bf 21}$ and ${\bf 18}$ points of $F$- score, respectively, for the language pair  Hindi-Tamil. Hindi and Tamil come from different language families, Indo Aryan and Dravidian, respectively, significantly boosting the efficacy of the cognate detection framework. On average, our supervised and weakly-supervised system has improved ${\bf 16.8}$ and ${\bf 9}$ points of $F$-score, respectively, on Indian language pairs.

Table \ref{table:2} shows the model performance on South African and Celtic language pairs. For South African languages, we have Zulu (zu), Swati (ss) and Xhosa (xh). Irish (ga), Manx (gv), and Scottish Gaelic (gd) represent the Celtic languages. We transferred morphological knowledge of Zulu and Irish to other South African and Celtic languages, respectively. Our proposed supervised and weakly-supervised framework outperformed the state-of-the-art baseline models. We reported poor performance of the baseline models for the language pairs Zulu and Swati. Due to the lack of training data for Zulu-Swati (only $23$ cognate pairs are available), the models cannot be trained effectively. On the other hand, our proposed approaches performed better than the baseline models by a large margin and very interestingly, our weakly-supervised model is better than the supervised model by ${\bf 6}$ points of $F$-score. The relatively complex word pairs such as \enquote{umgqibelo (Saturday)} in Zulu and \enquote{úm-gcibélo (Saturday)} in Swati are correctly identified as cognate pairs.

These results show that the proposed cognate detection framework can efficiently detect cognates across language pairs with the morphological knowledge of the pivot language. Moreover, with little training data, both the proposed weakly-supervised and unsupervised frameworks are an efficient solution for cognate detection.

\begin{table}[t]
\centering
\scriptsize
\begin{tabular}{lp{0.25cm}p{0.3cm}p{0.3cm}p{0.3cm}p{0.3cm}}
\toprule
{\textbf{Approaches / Languages}} & {\textbf{Mr-Gu}} & {\textbf{Pa-Gu}} & {\textbf{Mr-Pa}} & {\textbf{Mr-Bn}} & {\textbf{Gd-Gv}} \\ \midrule

\midrule
{Orthographic Similarity}    & {0.22}          & {0.26}          & {0.21}          & {0.32}          & {0.19}          \\
\cite{DBLP:conf/coling/Rama16} & {0.64} & {0.65}          & {0.69} & {0.61} & {0.49} \\ 
\cite{kanojia2019utilizing}                     & {0.72}          & {0.75}          & {0.77}          & {0.68}          & {0.64}         \\ \midrule
\multicolumn{6}{c}{\textbf{Proposed Supervised Methods}} \\
\hline
\midrule
{\em Proposed-method}                & {0.79}          & {0.79}          & {0.78}          & {0.74}          & {0.62}         \\ 
{\em $\text{Proposed-method}_\text{withknowledge}$}                & \textbf{{0.91}}          & \textbf{{0.88}}          & \textbf{{0.87}}          & \textbf{{0.86}}           & \textbf{{0.75}}                   \\ \midrule
\multicolumn{6}{c}{\textbf{Proposed Weakly-supervised/Unsupervised methods}} \\
\hline
\midrule
{\em $\text{Proposed-method}_\text{unspv}$}                 & {0.72} & {0.71} & {0.74} & {0.70} & {0.59} \\ 
{\em $\text{Proposed-method}_\text{wklysupv}$}              & {\textbf{0.86}} & {\textbf{0.83}} & {\textbf{0.84}} & {\textbf{0.80}} & {\textbf{0.73}} \\ \bottomrule
\end{tabular}
\caption{Results of supervised and weakly-unsupervised cognate detection task based on $\text{F}$-Score for Indian and Celtic languages in the absence of Pivot Languages.}
\label{table:3}
\end{table}

\subsection{Absence of Pivot Language}
\label{ssec:apl}

We now evaluate the robustness of our transfer learning approach on the cross-lingual language pairs in the absence of the pivot language. For Indian language pairs, our pivot language is Hindi (Hi), whereas for the Celtic language pairs, the pivot language is Irish (Ga). Table \ref{table:3}, shows that the transfer learning approach is still very efficient when the pivot language is absent. As we can see, for the language pairs Gd-Gv, without knowledge transfer in supervised learning, the recurrent neural network approach \citet{kanojia2019utilizing} is better than our approach. However, with the morphological knowledge encoded for both supervised and weakly-supervised methods, our model outperforms by ${\bf 11}$ and ${\bf 9}$ points of $F$-score, respectively. On average, our transferred knowledge-based weakly-supervised method has outperformed the baseline method by ${\bf 8.6}$ points of $F$-score.  Thus, we can see a steady performance across all language pairs, showing the stability of the proposed morphological knowledge transfer supervised and weakly-supervised framework.

\begin{table}[t]
\centering
\scriptsize
\begin{tabular}{lp{0.25cm}p{0.3cm}p{0.3cm}}
\toprule
{\textbf{Approaches / Languages}} & {\textbf{Hi-Mr}} & {\textbf{Hi-Pa}} & {\textbf{Mr-Bn}} \\ \midrule
\multicolumn{4}{c}{\textbf{Proposed Supervised Methods}} \\
\hline
\midrule
{\em With Hindi Knowledge}                & {0.91}          & {0.88}          & {0.86}  \\ 
{\em With Sanskrit Knowledge}                & {0.90}          & {0.86}         & {0.87} \\ \midrule
\multicolumn{4}{c}{\textbf{Proposed Weakly-supervised methods}} \\
\hline
\midrule
{\em With Hindi Knowledge}                 & {0.85} & {0.81} & {0.80} \\ 
{\em With Sanskrit Knowledge}                & {0.83} & {0.80} & {0.82}  \\ \bottomrule
\end{tabular}
\caption{Results of supervised and weakly-unsupervised cognate detection task based on $\text{F}$-Score for Indian languages transferring knowledge from Hindi and Sanskrit.}
\label{table:4}
\end{table}

\subsection{Knowledge of Historical Languages}
\label{ssec:hl}
In this section, we will discuss the effect of transferring knowledge from the historical language Sanskrit. Sanskrit is the historical ancestor of almost all the Indo-Aryan languages, thus making it one of the potential pivot languages to transfer the knowledge for the cognate detection task. 
We have studied the model's efficiency while transferring the knowledge of Sanskrit to modern languages. In this experiment, we have taken the Indian language pairs Hindi-Punjabi, Hindi-Marathi and Marathi-Bengali. Comparing the results of the models given in Table \ref{table:4}, we can observe a slight dip in model efficacy in both supervised and weakly-supervised frameworks while transferring the knowledge from Sanskrit compared to transferring the knowledge from Hindi to the language pairs Hindi-Punjabi and Hindi-Marathi. However, the performance for the language pair Marathi-Bengali has improved ${\bf 1.5}$ points in $F$-score on average.

We believe its performance can be attributed to the closeness and preserving more similarities in the characteristics of the language pairs Marathi and Bengali to Sanskrit than Hindi. Sanskrit is considered a highly agglutinating and morphologically rich language \cite{chatterji1926origin}; thus, it is hard to parse it computationally. Though Marathi and Bengali are not as morphologically complex as Sanskrit, the languages in this pair are more agglutinating and morphologically richer than Hindi and Punjabi. 

\subsection{Statistical Significance}

In this work, we  hypothesise that transferring morphological knowledge of the pivot language to the closely-related languages helps to identify cognates in both supervised and weakly-supervised settings. To compute the performance of each language-pairs from Table \ref{table:1}, \ref{table:2}, \ref{table:3} we run the models on two different settings and obtain the distribution of the performance scores: (i) we have run $5$-fold cross-validations two times (which makes a total of $10$ sets of results), and (ii) we kept $1$-fold for a single test set and ran it $10$ times for $10$ different sets of results. Two sample t-tests showed that our results are statistically significant in both the cases over the baseline models ($p < 0.01$).

\section{Ablation Study}
\label{sec:ablat}

\begin{figure}[t]
    \centering
    \hspace*{-6mm}\includegraphics[width=0.95\columnwidth]{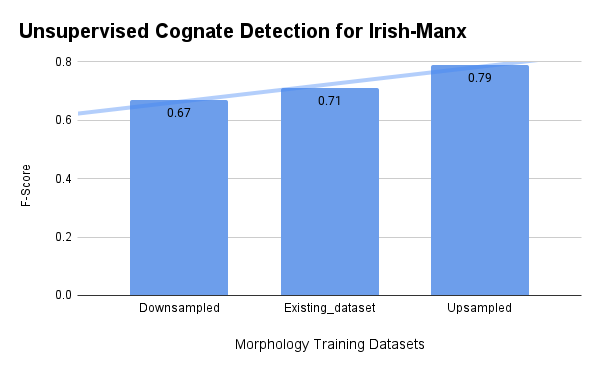}
    \caption{$\text{F}$-Score for Irish-Manx language pairs transferring knowledge from different morphological learners}
    \label{fig:weight-parameter-selction}
\end{figure}


\noindent{\bf Effect of morphological training datasize}. One of the challenges of morphological knowledge transfer is the efficient learning of word structure in the presence of a few morphological training datasets. As Irish has few training datasets, we have evaluated the proposed framework on different samples of the Irish-Manx dataset (Refer to Section \ref{sec:exper}). We have down-sampled and up-sampled the morphology training set to $30\%$ compared to the existing datasets. From Figure \ref{fig:weight-parameter-selction}, we can observe that the best $F$-score ${\bf 0.79}$ is achieved when the morphological learner model is trained with $30\%$ more data size than the original size. This emphasizes our claim of the model's efficacy even with a slight increase in the morphological training datasets, which opens up the opportunity of implementing the weakly-supervised cognate detection framework on diverse under-resourced languages.

\section{Conclusion}

This paper proposed a novel {\em language agnostic weakly-supervised cognate detection framework based on Siamese architecture}.
Experiments on three different datasets consisting of Indian languages, Celtic languages and South-African languages showcase the efficacy of our framework in understanding the structural relations between cross-lingual words across languages. We also show that transferring morphological knowledge to closely-related word pairs with the help of a shared encoder improves the model's efficacy in different scenarios. Our study on knowledge transfer from historical languages depicts changes in the word structures of modern languages. We demonstrate that our approach outperforms the existing supervised and semi-supervised frameworks and establishes state-of-the-art results for the cognate detection task. We also showcase the stability of learning morphology on a small number of training datasets, which opens up the possibility of deploying the system across language families.

Our future work will design and compare a semi-supervised framework based on labelled and unlabelled training sets. We will study whether the semi-supervised framework improves efficiency while detecting the cognates. 

\section*{Limitations}

During the evaluation, we have not experimented with choosing multiple languages as pivot languages in the same language family. So, the performance of the transfer learning framework may change depending on the choice of the pivot language. Also, during the evaluation, we mostly conducted our experiments on modern language pairs. Thus, the performance of the framework may differ for studies of historical linguistic. From the training perspective, more fine-tuning may improve the performance of the models but we have compared the results produced with the settings described in our work.

\section*{Acknowledgements}

This publication was supported by a research grant from Irish Research Council: Grant IRCLA/2017/129 (CARDAMOM-Comparative Deep Models of Language for Minority and Historical Languages) co-funded by Science Foundation Ireland (SFI) under Grant SFI/18/CRT/6223 (CRT-Centre for Research Training in Artificial Intelligence) and
Science Foundation Ireland (SFI) under Grant SFI/12/RC/2289$\_$P2 (Insight$\_$2).

\bibliography{anthology,custom}
\bibliographystyle{acl_natbib}

\appendix



\end{document}